\documentclass{IJCAS}

\usepackage{url}
\usepackage{array,tabularx}
\usepackage{multicol} 
\usepackage{bm}
\usepackage{multirow}

\journalvolumn{VV}
\journalnumber{X}
\journalyear{YYYY}
\setarticlestartpagenumber{1}

\allowdisplaybreaks

\begin{document}
\title{Phase-based Nonlinear Model Predictive Control for Humanoid Walking Stabilization with Single and Double Support Time Adjustments}

\author{
Kwanwoo Lee\orcid{0009-0001-7897-4310}, 
Gyeongjae Park\orcid{0000-0001-6429-1036}, 
Myeong-Ju Kim\orcid{0000-0002-9955-4307},
and Jaeheung Park*\orcid{0000-0002-5062-8264}
}

\begin{abstract}
The contact sequence of humanoid walking consists of single and double support phases (SSP and DSP), and their coordination through proper duration and dynamic transition based on the robot’s state is crucial for maintaining walking stability. 
Numerous studies have investigated phase duration optimization as an effective means of improving walking stability.
This paper presents a phase-based Nonlinear Model Predictive Control (NMPC) framework that jointly optimizes Zero Moment Point (ZMP) modulation, step location, SSP duration (step timing), and DSP duration within a single formulation.
Specifically, the proposed framework reformulates the nonlinear DCM~(Divergent Component of Motion) error dynamics into a phase-consistent representation and incorporates them as dynamic constraints within the NMPC.
The proposed framework also guarantees ZMP input continuity during contact-phase transitions and disables footstep updates during the DSP, thereby enabling dynamically reliable balancing control regardless of whether the robot is in SSP or DSP.
The effectiveness of the proposed method is validated through extensive simulation and hardware experiments, demonstrating improved balance performance under external disturbances.
\end{abstract}


\maketitle

\makeAuthorInformation{
Manuscript received January 10, 2025; revised March 10, 2025; accepted May 10, 2025. Recommended by Associate Editor Soon-Shin Lee under the direction of Editor Milton John.\\

Kwanwoo Lee, Gyeongjae Park, and Jaeheung Park are with the Department of Intelligence and Information, Seoul National University, Republic of Korea (e-mails: \{kwlee365, rudwo1301, park73\}@snu.ac.kr). 
Myeong-Ju Kim is with the Robotics Lab, Hyundai Motor Group, Republic of Korea (e-mail: myeong-ju@snu.ac.kr). 
Jaeheung Park is also with the Advanced Institutes of Convergence Technology, Republic of Korea, and with ASRI, AIIS, Seoul National University, Republic of Korea. 
He is the corresponding author of this paper (e-mail: park73@snu.ac.kr).

* Corresponding author.
}

\runningtitle{2025}{Kwanwoo Lee, Gyeongjae Park, Myeong-Ju Kim, and Jaeheung Park}{Manuscript Template for the International Journal of Control, Automation, and Systems: ICROS {\&} KIEE}{xxx}{xxxx}{x}

\section{INTRODUCTION}
\label{sec1}
Balance control is essential for humanoid robots to perform stable locomotion and tasks in real-world environments with various uncertainties.
However, achieving robust balance control is challenging due to the complex dynamics of humanoid robots, characterized by nonlinearity, high degrees of freedom, and an underactuated floating base, as well as external disturbances caused by uneven terrain and potential collisions.

Accordingly, optimization-based control for legged locomotion using Quadratic Programming (QP)~\cite{kuindersma2016optimization, khadiv2020walking, jeong2019robust, shafiee2019online, griffin2017walking} or Model Predictive Control (MPC)~\cite{ding2022oampc, choe2023seamless, scianca2020stability, hong2020real, kim2025real, kim2025tro} has been actively studied as a key class of approaches for humanoid balance control.
Rather than directly utilizing the complex dynamics of humanoid robots, these approaches commonly rely on reduced-order models to simplify the control problem, including the Linear Inverted Pendulum Model (LIPM)~\cite{kajita20013d}, the Single Rigid Body Model (SRBM)~\cite{carlo2018srbmpc}, and the Divergent Component of Motion (DCM)~\cite{takenaka2009real, englsberger2015three}. 
These methods typically formulate legged locomotion as an optimization problem, where decision variables such as contact wrenches (or the Zero Moment Point~(ZMP)~\cite{vukobratovic2004zero}), step location, and step timing (i.e., the Single Support Phase (SSP) duration) are optimized. 
By solving such optimization problems online, the resulting control solutions can be used to enable humanoid robots to maintain balance over uneven terrain and mitigate the effects of external disturbances during locomotion.
For example, Ding \textit{et al}.~\cite{ding2022oampc} proposed a linear MPC~(LMPC) framework based on the SRBM that incorporates step location adjustment for dynamic humanoid walking. 
Choe \textit{et al}.~\cite{choe2023seamless} presented a nonlinear MPC~(NMPC) framework derived from the LIPM that simultaneously optimizes ZMP modulation, step location, step timing, and the time derivative of the centroidal angular momentum.
Although these studies have demonstrated effective balance-control performance, many existing methods still do not explicitly incorporate the Double Support Phase~(DSP) into the optimization problem, often assuming a fixed DSP duration or omitting the DSP entirely from the gait cycle.

The importance of DSP in the nominal gait schedule has been emphasized to enable proper weight shifting to unload the feet~\cite{griffin2023reachability}.
Moreover, the absence of DSP can lead to discontinuous variations in robot CoM acceleration during support transitions, resulting in discontinuous joint torque commands~\cite{englsberger2015three}.
Although this may not be problematic for robots with high-bandwidth torque actuators, it can pose significant implementation challenges for most robots lacking such high-performance actuators.
Thus, DSP remains important for achieving feasible and smooth walking in such humanoid robotic systems.
However, when multiple steps are required instantaneously to maintain balance, a fixed DSP duration can delay the transition to the subsequent SSP.
Therefore, several studies have employed heuristics to optimize the DSP duration~\cite{griffin2023reachability, kim2023foot, egle2023step}.
Kim~\textit{et~al.}~\cite{kim2023foot} proposed a DSP scaling method that modifies the DSP duration according to updates in the planned footstep.
However, this approach adjusts the DSP duration only under specific timing conditions, making the controller sensitive to when disturbances occur and limiting its general applicability.
Egle~\textit{et~al.}~\cite{egle2023step} proposed a walking control framework that optimizes ZMP modulation, step location, both SSP and DSP durations using LMPC.
However, their formulation introduces simplifying assumptions on the DCM-ZMP dynamics, where ZMP modulation in the current phase and phase-duration adjustments in future phases are set to zero for linearization.
These assumptions require hand-crafted coordination mechanisms, such as DCM error separation and projection, to compensate for the resulting model mismatch.
Consequently, the successive linearization reduces model fidelity and often leads to motion errors compared with approaches that retain the full nonlinear dynamics.

In this paper, a robust balance control framework is proposed to overcome disturbances through the simultaneous optimization of ZMP modulation, step location, and both SSP and DSP durations.
The proposed framework establishes a unified control structure that integrates the aforementioned balance strategies, ensuring dynamically reliable balance control regardless of whether the robot is in SSP or DSP.

The main contributions of this study are as follows:
\begin{itemize}
    \item 
    This study proposes an NMPC framework that unifies the optimization-based balance control problem, including ZMP modulation, step location, and phase durations, in a phase-consistent manner.
    While prior DCM-based control frameworks~\cite{jeong2019robust, choe2023seamless} primarily support balance control only during the SSP, the proposed method extends DCM tracking control to both SSP and DSP within a single MPC formulation.

    \item 
    In contrast to~\cite{egle2023step}, the proposed framework unifies the aforementioned balance strategies within a single control structure while enforcing ZMP input continuity across contact-phase transitions and disabling footstep updates during the DSP.
    These design choices ensure dynamically consistent balance control and lead to improved performance compared with~\cite{egle2023step}.

    \item The effectiveness of the proposed framework is validated through extensive simulation and real-world experiments. 
    Compared with three baseline controllers~\cite{choe2023seamless, kim2023foot, egle2023step}, the proposed method demonstrates improved robustness under large external disturbances and on uneven terrains.

\end{itemize}

The remainder of this paper is organized as follows.
Section~\ref{sec:preliminaries} reviews the LIPM and DCM preliminaries.
Section~\ref{sec:overview} presents an overview of the proposed framework.
Section~\ref{sec:nmpc} formulates the phase-based NMPC, including its cost function and constraints.
Section~\ref{sec:wbc} introduces the whole-body controller used to track the NMPC outputs.
Section~\ref{sec:result} reports the simulation and experimental results.
Finally, Section~\ref{sec:conclusion} concludes the paper.

\section{Preliminaries} \label{sec:preliminaries}
\subsection{Linear inverted pendulum model}
The LIPM has been widely utilized as a reduced model to simplify the complex dynamics of a humanoid robot into a linear relationship between the CoM and the ZMP~\cite{kajita20013d}.
In the LIPM, the total mass of the robot is assumed to be concentrated at the CoM. 
Both the time derivative of the centroidal angular momentum and the vertical acceleration of the CoM are neglected.
Under these assumptions, the relationship between the CoM and the ZMP is given by
\begin{equation}
\ddot{\bm{c}}= \frac{g}{c_z - z_z}(\bm{c} - \bm{z}),
\label{lipm}
\end{equation}
where $\bm{c}\in\mathbb{R}^2$ and $\bm{z}\in\mathbb{R}^2$ denote the CoM and ZMP positions in the horizontal plane, respectively.  
The variables $c_z$ and $z_z$ represent the vertical positions of CoM and ZMP, respectively. 
The variable $g$ is the gravitational acceleration.

\subsection{Divergent component of motion}
The LIPM in~\eqref{lipm} contains both stable and unstable eigenmodes.
The DCM corresponds to the unstable mode associated with the positive eigenvalue. 
The relationship between DCM and CoM is defined as,
\begin{equation}
\bm{\xi} = \bm{c} + b\bm{\dot{c}},
\label{dcm-com}
\end{equation}
where the variable $\bm{\xi} \in \mathbb{R}^2$ denotes the DCM position and $b=\sqrt{\frac{c_z - z_z}{g}}$ is the time constant.
By reordering \eqref{dcm-com}, the relationship between the DCM and the ZMP is derived as follows:
\begin{equation}
\bm{\dot{\xi}} = \frac{1}{b} (\bm{\xi} - \bm{z}).
\label{dcm-zmp}
\end{equation}

\subsection{DCM-ZMP dynamics}
A gait schedule of a bipedal walking robot can be represented as a sequence of $n_{\varphi}$ (single and double) support phases.
During the SSP, the robot supports itself with a single foot while swinging the other foot toward the target step location.
During the DSP, both feet are in contact with the ground as the robot shifts its weight to the next support foot. 
For each phase $\varphi \in \{1, \cdots, n_{\varphi}\}$, the current ZMP position $\bm{z}_{\varphi,t}$ is interpolated in the horizontal plane from the starting point of ZMP $\bm{z}_{\varphi,0}$ to the end point of ZMP $\bm{z}_{\varphi,T}$.
In this study, the ZMP trajectory is planned using linear interpolation within each phase~$\varphi$:
\begin{equation}
\bm{z}_{\varphi,t}= \left(1-\frac{t}{T_\varphi}\right)\bm{z}_{\varphi,0}+\frac{t}{T_\varphi}\bm{z}_{\varphi,T}. 
\label{linzmp}
\end{equation}
Here, $t \in [0, T_\varphi]$ denotes the current time within the phase~$\varphi$, and $T_\varphi$ is the duration of phase $\varphi$.

Since the dynamics of DCM-ZMP~(\ref{dcm-zmp}) is a first-order differential equation and the ZMP trajectory in~(\ref{linzmp}) is linear, the solution to (\ref{dcm-zmp}) is obtained using partial integration as
\begin{equation}
\bm{\xi}_{\varphi,T} =\mathbf{Z}_\alpha(T_{\varphi}) + e^{\frac{T_{\varphi} - t}{b}} 
\left( \bm{\xi}_{\varphi,t} - \mathbf{Z}_\beta(T_{\varphi}, t) \right), \\
\label{sol dcm-zmp}
\end{equation}
where
\begin{equation}
\begin{split}
&\mathbf{Z}_\alpha(T_{\varphi}) = \bm{z}_{\varphi,T} + \frac{b}{T_{\varphi}} 
\left( \bm{z}_{\varphi,T} - \bm{z}_{\varphi,0} \right), \\
&\mathbf{Z}_\beta(T_{\varphi}, t) = \bm{z}_{\varphi,0} + \frac{t + b}{T_{\varphi}} 
\left( \bm{z}_{\varphi,T} - \bm{z}_{\varphi,0} \right).
\end{split}
\end{equation}
Please refer to~\cite{englsberger2017smooth} for the corresponding analytical solutions of the DCM–ZMP dynamics for various $n$th-order spline representations.
Hereafter, the subscript $\varphi$ is omitted for better readability.

\subsection{DCM offset}
\begin{figure}[!t]
\centerline{\includegraphics[width=0.495\textwidth]{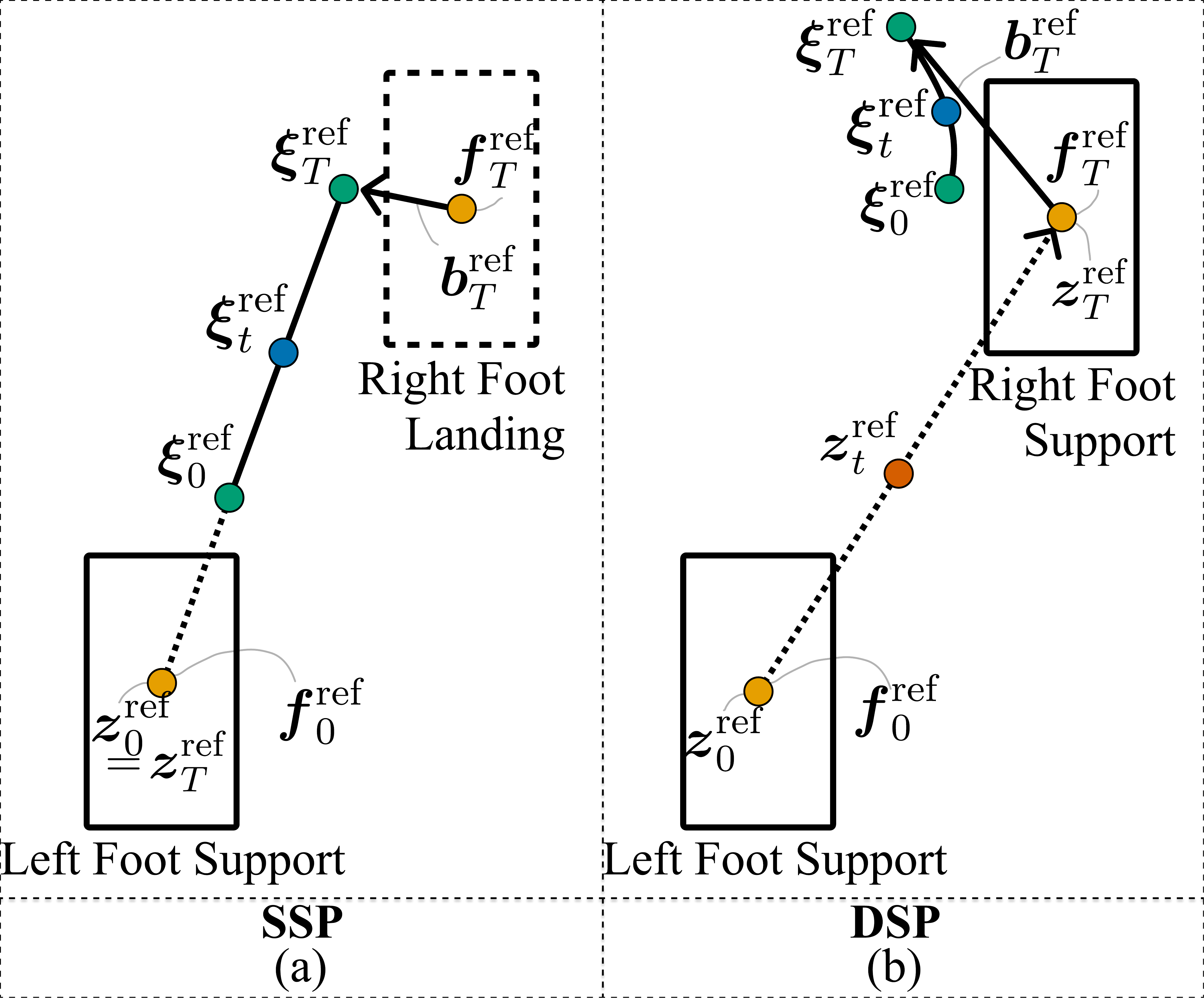}}
\caption{
Graphical illustration of the reference DCM trajectory across gait phases:
(a) SSP, where the ZMP is fixed at the center of the support foot and the reference DCM diverges toward the upcoming landing foot;
(b) DSP, where the ZMP trajectory transitions linearly between both feet, enabling a smooth and continuous evolution of the reference DCM.
}
\label{fig/endpoint}
\end{figure}
The DCM end point $\bm{\xi}_T$ represents the position of the DCM at the end of each phase. 
Fig.~\ref{fig/endpoint} illustrates a graphical diagram of the reference DCM trajectory. 
During SSP, $\bm{f}_0$ and $\bm{f}_T$ represent the current position of the support foot and the target step location of the swing foot, respectively.
The DCM end point $\bm{\xi}_T$ during the SSP can be decomposed into the step location $\bm{f}_T$ and a DCM offset $\bm{b}_T$ as follows~\cite{khadiv2020walking}:
\begin{equation}
\bm{\xi}_T = \bm{f}_T + \bm{b}_T.
\label{dcm-offset}
\end{equation}
According to the definition of the DCM, if the robot steps onto the DCM within the $N$-step capture region, it will eventually come to a complete stop~\cite{koolen2012capturability}.
This observation implies that the reference DCM offset $\bm{b}^{\text{ref}}_T$ should be determined from the reference step location $\bm{f}^{\text{ref}}_T$ to maintain the desired walking velocity.
Therefore, minimizing the DCM offset error $\bm{b}_T^{\text{err}}$ contributes to maintaining the target walking velocity and preventing the robot from falling due to disturbances.

\begin{figure*}[!t]
\centerline{\includegraphics[width=\textwidth]{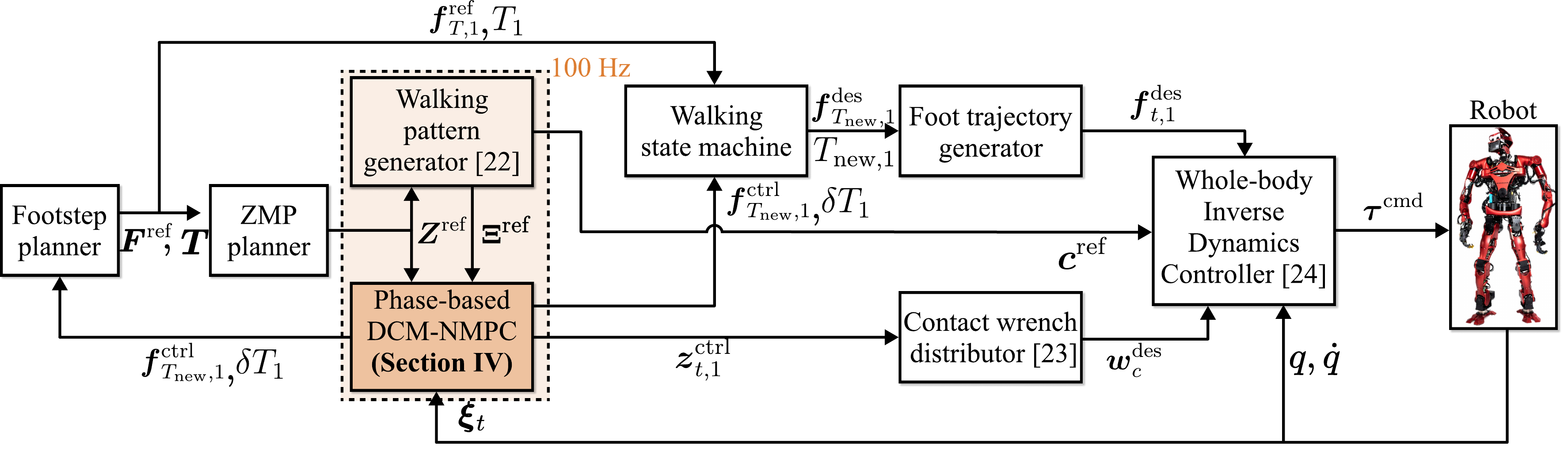}}
\caption{Overall control framework for humanoid balance control utilizing the proposed phase-based Nonlinear Model Predictive Control.}
\label{fig/blockdiagram}
\end{figure*}

In this paper, we further extend the definition of DCM offset to the DSP. The nominal gait schedule consists of a repeated cycle that begins with the SSP and ends with the DSP.
In an attempt to apply the relationship in \eqref{dcm-offset} to the DSP as well, we define $\bm{f}_0$ and $\bm{f}_T$ in the DSP as the support foot position and the swing foot landing position from the previous SSP, respectively. 
By definition of the DSP, both $\bm{f}_0$ and $\bm{f}_T$ remain unchanged during the DSP.
Therefore, (\ref{dcm-offset}) is generally applied in all phases without loss of generality.
This allows the structure of the optimization problem and the variables to remain consistent in all phases.

\section{Overall Control Framework} \label{sec:overview}
This section presents the overall control framework for humanoid balance control, which includes the walking pattern generator, the phase-based NMPC, and the QP-based whole-body controller.
Fig.~\ref{fig/blockdiagram} illustrates the overall control architecture.
First, the footstep planner generates the reference footstep locations~$\bm{F}^{\text{ref}}$ and the phase durations~$\bm{T}$.
Using these references, the reference ZMP trajectory is planned such that it remains in the center of the support foot during SSP (i.e., $\bm{z}^{\text{ref}}_{0} = \bm{z}^{\text{ref}}_{T}$) and is linearly interpolated between the centers of the previous and next support feet during DSP.
The walking pattern generator in~\cite{wieber2006trajectory} produces the reference CoM position and velocity trajectories $\bm{C}^{\text{ref}}, \dot{\bm{C}}^{\text{ref}} \in \mathbb{R}^{N_p \times 2}$ to track the reference ZMP trajectory $\bm{Z}^{\text{ref}} \in \mathbb{R}^{N_p \times 2}$ while minimizing the CoM jerk within a preview horizon of~$N_p$.
Then, the DCM trajectory $\bm{\Xi}^{\text{ref}} \in \mathbb{R}^{N_p \times 2}$ is generated using~(\ref{dcm-com}).
Note that all CoM, ZMP, and DCM trajectories are defined in the horizontal plane.

The phase-based NMPC optimizes the decision variable 
$\bm{v}_i = \begin{bmatrix}
{\bm{z}^{\text{ctrl}}_{0,i}}^\top &
{\bm{z}^{\text{ctrl}}_{T,i}}^\top &
{\bm{f}^{\text{ctrl}}_{T_{\text{new}},i}}^\top &
{\bm{b}^{\text{err}}_{T_{\text{new}},i}}^\top &
\delta T_{i}
\end{bmatrix}^\top \in \mathbb{R}^9$
for each phase $i = 1, 2, \cdots, n_c$, where $n_c$ denotes the number of previewed phases.
The superscript $\text{ctrl}$ and $\text{err}$ denote control and error terms, respectively.
The proposed NMPC receives the current DCM measurement $\bm{\xi}_t \in \mathbb{R}^2$, together with the reference ZMP and DCM trajectories, $\bm{Z}^{\text{ref}}$ and $\bm{\Xi}^{\text{ref}}$, and outputs the ZMP modulation terms ${\bm{z}^{\text{ctrl}}_{0,i}}, {\bm{z}^{\text{ctrl}}_{T,i}} \in \mathbb{R}^2$, the step location adjustment $\bm{f}^{\text{ctrl}}_{T_{\text{new}},i} \in \mathbb{R}^2$, and the phase duration adjustment $\delta T_i$ to minimize the DCM offset error $\bm{b}^{\text{err}}_{T_{\text{new}},i}\in \mathbb{R}^2$.

The ZMP control input $\bm{z}_{t,1}^{\text{ctrl}}$ is linearly interpolated from the start point $\bm{z}_{0,1}^{\text{ctrl}}$ to the end point $\bm{z}_{T,1}^{\text{ctrl}}$ over the optimized phase duration $T_{\text{new},1}$ according to~(\ref{linzmp}). 
The ZMP distributor~\cite{kajita2010biped} converts the optimized target ZMP, defined as $\bm{z}_{t,1}^{\text{des}} = \bm{z}_{t,1}^{\text{ref}} + \bm{z}_{t,1}^{\text{ctrl}}$, into the desired contact wrenches $\bm{w}_c^{\text{des}} \in \mathbb{R}^{12}$ applied to both feet.

The target step location of the swing foot $\bm{f}_{T_{\text{new}},1}^{\text{des}}$ is determined by adding the optimized step location adjustment $\bm{f}^{\text{ctrl}}_{T_{\text{new}},1}$ to the reference landing position $\bm{f}_{T,1}^{\text{ref}}$.  
Similarly, the optimized phase duration $T_{\text{new},1}$ is computed by adding the phase duration adjustment $\delta T_1$ to the reference duration $T_1$.
The swing foot trajectory $\bm{f}_{t,1}^{\text{des}}$ is then generated using a fifth-order spline.
When the current time $t$ reaches the optimized phase duration $T_{\text{new},1}$, a phase transition is triggered; then, $t$ is reset to zero, and the system progresses to the next phase.

The joint control module computes the desired angles of the lower body joints $\bm{q}_{lb}^{\text{des}}$ using analytical inverse kinematics based on the optimized trajectories of the feet and CoM.
The desired angles of the upper body joints $\bm{q}_{ub}^{\text{des}}$ remain constant.
The joint torque command $\bm{\tau}^{\text{cmd}}$ is computed as the sum of the feedforward torque and the PD feedback torque.  
Here, the feedforward torque $\bm{\tau}^{\text{ff}}$ is computed by the QP-based whole-body controller~(WBC) from~\cite{kim2020dynamicwbc} described in Section~\ref{sec:wbc}.
The WBC aims to track the desired joint accelerations $\bm{\ddot{q}}^{\text{des}}$ and the contact wrenches $\bm{w}_c^{\text{des}}$.  
The PD feedback torque $\bm{\tau}^{\text{pd}}$ is calculated in joint space to track the desired joint positions.

\section{Phase-based Nonlinear Model Predictive Control}\label{sec:nmpc} 
This section introduces the proposed phase-based NMPC framework. The formulation is presented in three parts: the derivation of the nonlinear DCM error dynamics based on the DCM–ZMP relationship in~\eqref{sol dcm-zmp} under phase duration adjustment, the definition of the optimization variables, and the resulting NMPC problem that integrates these components.


\subsection{Nonlinear DCM error dynamics} 
The ZMP start point $\bm{z}_0$, ZMP end point $\bm{z}_T$, current DCM $\bm{\xi}_t$, and DCM end point $\bm{\xi}_T$ can be decomposed as reference and error dynamics.
Hence,~\eqref{sol dcm-zmp} can be split into the reference and error dynamics of the DCM as follows:
\begin{equation}
\bm{\xi}_{T}^{\text{ref}} = \mathbf{Z}^{\text{ref}}_\alpha(T) + e^{\frac{T -t}{b}} \left( \bm{\xi}^{\text{ref}}_{t} - \mathbf{Z}^{\text{ref}}_\beta(T, t) \right),
\label{sol dcm-zmp ref}
\end{equation}
\begin{equation}
\bm{\xi}_{T}^{\text{err}} = \mathbf{Z}^{\text{err}}_\alpha(T) + e^{\frac{T - t}{b}} \left( \bm{\xi}^{\text{err}}_{t} - \mathbf{Z}^{\text{err}}_\beta(T, t) \right).
\label{sol dcm-zmp err}
\end{equation} 

Phase duration adjustment is described as the process of finding an optimized duration $T_{\text{new}} = T + \delta T$ that aligns the reference DCM trajectory $\bm{\xi}_{t_{\text{new}}}^{\text{ref}}$ more closely with the measured DCM $\bm{\xi}_t$~\cite{griffin2023reachability}. 
As noted in~\cite{jeong2019robust}, the updated DCM reference $\bm{\xi}^{\text{ref}}_{t_{\text{new}}}$ can be guided to converge toward the originally planned end point by ensuring that the reference end point remains unchanged during phase duration adjustment (i.e., $\bm{\xi}_T^{\text{ref}} = \bm{\xi}_{T_{\text{new}}}^{\text{ref}}$).
This also helps the updated DCM reference better align with the current DCM measurement $\bm{\xi}_t$.
Equations~(\ref{sol dcm-zmp ref}) and~(\ref{sol dcm-zmp err}) remain valid under the new phase duration and can be rewritten as:
\begin{equation}
\bm{\xi}_{T_{\text{new}}}^{\text{ref}} = \mathbf{Z}^{\text{ref}}_\alpha(T_{\text{new}}) + e^{\frac{T_{\text{new}} - t}{b}} \left( \bm{\xi}^{\text{ref}}_{t_{\text{new}}} - \mathbf{Z}^{\text{ref}}_\beta(T_{\text{new}}, t) \right),
\label{sol dcm-zmp ref new}
\end{equation}
\begin{equation}
\bm{\xi}_{T_{\text{new}}}^{\text{err}} = \mathbf{Z}^{\text{ctrl}}_\alpha(T_{\text{new}}) + e^{\frac{T_{\text{new}} - t}{b}} \left( \bm{\xi}^{\text{err}}_{t_{\text{new}}} - \mathbf{Z}^{\text{ctrl}}_\beta(T_{\text{new}}, t) \right).
\label{sol dcm-zmp err new}
\end{equation}
Using the condition $\bm{\xi}^{\text{ref}}_{T} = \bm{\xi}^{\text{ref}}_{T_{\text{new}}}$, (\ref{sol dcm-zmp ref}) and~(\ref{sol dcm-zmp ref new}) yield the relationship between the current and updated reference DCMs as follows:
\begin{equation}
\begin{split}
&e^{\frac{T_{\text{new}} - t}{b}} \left( \bm{\xi}_{t_{\text{new}}}^{\text{ref}} - \bm{Z}_{\beta}^{\text{ref}}(T_{\text{new}}, t) \right)
\\&=  \bm{Z}_{\alpha}^{\text{ref}}(T) - \bm{Z}_{\alpha}^{\text{ref}}(T_{\text{new}}) + e^{\frac{T - t}{b}} \left( \bm{\xi}_t^{\text{ref}} - \bm{Z}_{\beta}^{\text{ref}}(T, t) \right).
\end{split}
\label{relationship between the current and new reference DCM}
\end{equation}
Substituting $\bm{\xi}^{\text{err}}_{t_{\text{new}}} = \bm{\xi}_t - \bm{\xi}^{\text{ref}}_{t_{\text{new}}}$ into~(\ref{sol dcm-zmp err new}), and using~(\ref{relationship between the current and new reference DCM}), the DCM error dynamics for the new phase duration can be expressed as:
\begin{equation}
\begin{split}
&\bm{\xi}_{T_{\text{new}}}^{\text{err}} \\
&= \bm{Z}_{\alpha}^{\text{ctrl}}(T_{\text{new}})
+ e^{\frac{T_{\text{new}} - t}{b}} \left( \bm{\xi}_t - \bm{Z}_{\beta}(T_{\text{new}}, t) \right)\\
&- \left( \bm{Z}_{\alpha}^{\text{ref}}(T) - \bm{Z}_{\alpha}^{\text{ref}}(T_{\text{new}}) + e^{\frac{T - t}{b}} \left( \bm{\xi}_t^{\text{ref}} - \bm{Z}_{\beta}^{\text{ref}}(T, t) \right) \right).
\end{split}
\end{equation}
By decomposing the DCM end-point error as $\bm{\xi}_{T_{\text{new}}}^{\text{err}} = \bm{f}_{T_{\text{new}}}^{\text{ctrl}} + \bm{b}_{T_{\text{new}}}^{\text{err}}$, and dividing the current DCM $\bm{\xi}_t$ into reference and error parts, the nonlinear DCM error dynamics for the new phase duration $T_{\text{new}}$ can be reformulated as follows:
\begin{equation}
\begin{split}
&\bm{F}_i(\bm{v}_i, \bm{\xi}^{\text{err}}_{t,i}, t) \\
& =
\bm{f}^{\text{ctrl}}_{T_{\text{new}}} + \bm{b}^{\text{err}}_{T_{\text{new}}}
- \bm{Z}_\alpha^{\text{ctrl}}(T_{\text{new}})
- \bm{Z}_\alpha^{\text{ref}}(T_{\text{new}}) \\
& + \bm{Z}_\alpha^{\text{ref}}(T) 
 - e^{\frac{T_{\text{new}} - t}{b}} 
\left( \bm{\xi}_t^{\text{err}} - \bm{Z}_\beta^{\text{ctrl}}(T_{\text{new}}, t) \right) \\
& - \left( e^{\frac{T_{\text{new}} - t}{b}} - e^{\frac{T - t}{b}} \right)
\bm{\xi}_t^{\text{ref}} \\
& + e^{\frac{T_{\text{new}} - t}{b}} \bm{Z}_\beta^{\text{ref}}(T_{\text{new}}, t)
- e^{\frac{T - t}{b}} \bm{Z}_\beta^{\text{ref}}(T, t)
= \bm{0}.
\end{split}
\label{eq/optdyn}
\end{equation}
To summarize, \eqref{eq/optdyn} is derived under the assumption that the reference end point of the DCM remains unchanged despite the phase duration adjustment.
The objective is to better align the updated DCM trajectory with the current DCM measurement.
It serves as a nonlinear constraint to minimize the DCM offset error $\bm{b}^{\text{err}}_{T_{\text{new}}}$ by optimizing the ZMP modulation~$\bm{z}^{\text{ctrl}}_0$, $\bm{z}^{\text{ctrl}}_T$, step location adjustment~$\bm{f}^{\text{ctrl}}_{T_{\text{new}}}$, and phase duration adjustment $\delta T$.  

\subsection{Optimization variables} \label{sec:opt_var}
When the number of previewed phases is given by $n_c$, the optimization variable $\bm{v}_i \in \mathbb{R}^9$ is defined as follows:
\begin{equation}
 \bm{v}_i = \begin{bmatrix}
{\bm{z}^{\text{ctrl}}_{0,i}}^\top&
{\bm{z}^{\text{ctrl}}_{T,i}}^\top&
{\bm{f}^{\text{ctrl}}_{T_{\text{new}},i}}^\top&
{\bm{b}^{\text{err}}_{T_{\text{new}},i}}^\top&
 \delta T_{i} 
\end{bmatrix}^\top,
\forall i
\label{eq/optvar}
\end{equation}
Here, $i = 1$ corresponds to the current support phase of the robot, while $i = 2, \dots, n_c$ represent the previewed phases within the MPC horizon.

\subsection{Nonlinear MPC formulation}
With the optimization variables $\bm{v} = 
\begin{bmatrix}
\bm{v}_1^\top & \cdots & \bm{v}_{n_c}^\top
\end{bmatrix}^\top \in \mathbb{R}^{9n_c}$, the phase-based NMPC is formulated as follows:
\begin{align}
\min_{\bm{v}} \quad & \sum_{i=1}^{n_c} \bm{v}_i^\top \bm{W}_{\bm{v}_i} \bm{v}_i \label{eq/cost}
\\
\text{s.t.} \quad 
&\begin{aligned}
& \bm{F}_1(\bm{v}_1, \bm{\xi}_{t,1}^{\text{err}}, t) = \bm{0}   
\label{eq/ceq1}
\\
& \bm{F}_i(\bm{v}_i, \bm{\xi}_{0,i}^{\text{err}}, 0) = \bm{0} \quad \text{for } i = 2, \ldots, n_c 
\end{aligned}
\\
& \bm{z}_{T,i}^{\text{ctrl}} = \bm{z}_{0,i+1}^{\text{ctrl}} \quad \text{for } i = 1, \ldots, n_c - 1 
\label{eq/ceq2}
\\
&\begin{aligned}
    & \underline{\bm{z}}^{\text{ctrl}}_{0,i} \leq \bm{z}^{\text{ctrl}}_{0,i} \leq \overline{\bm{z}}^{\text{ctrl}}_{0,i} \\ 
    &\underline{\bm{z}}^{\text{ctrl}}_{T,i} \leq \bm{z}^{\text{ctrl}}_{T,i} \leq \overline{\bm{z}}^{\text{ctrl}}_{T,i}
\end{aligned}
\quad \forall i
\label{eq/cineq1}
\\ 
&\begin{aligned}
    & \underline{\bm{f}}^{\text{ctrl}}_{T_\text{new},i,j} - M(\bm{1} - \bm{y}_j) 
    \leq \bm{f}^{\text{ctrl}}_{T_\text{new},i}  
    \\
    & \bm{f}^{\text{ctrl}}_{T_\text{new},i}  
    \leq \overline{\bm{f}}^{\text{ctrl}}_{T_\text{new},i,j} + M(\bm{1} - \bm{y}_j)
    \quad  
    \\ 
    & \forall i \quad \forall j \in \{\text{SSP}, \text{DSP} \} 
\end{aligned}
\label{eq/cineq2}
\\
& \underline{\delta T}_i \leq \delta T_i \leq \overline{\delta T}_i 
\quad \forall i.
\label{eq/cineq3}
\end{align}
Here, $\bm{W}_{\bm{v}_i}$ denotes a set of positive definite weighting matrices.
The underline and overline indicate the lower and upper bounds of the corresponding variable, respectively.

\begin{table}[t]
\centering
\caption{Weighting parameters for the proposed NMPC.}
\label{tab:weights}
\renewcommand{\arraystretch}{1.2}
\setlength{\tabcolsep}{6pt}
\begin{tabular}{ccccc}
\toprule
\multicolumn{5}{c}{$\bm{W}_{\bm{v}_i}$} \\
\midrule
$\bm{z}^{\text{ctrl}}_{0,i}$ &
$\bm{z}^{\text{ctrl}}_{T,i}$ &
$\bm{f}^{\text{ctrl}}_{T_{\text{new},i}}$ &
$\bm{b}^{\text{err}}_{T_{\text{new},i}}$ &
$\delta T_{i}$ \\
\midrule
$[1, 1]^{\top}$ &
$[1, 1]^{\top}$ &
$[0.01, 0.01]^{\top}$ &
$[500, 500]^{\top}$ &
$100$ \\
\bottomrule
\end{tabular}
\end{table}
The cost function in~\eqref{eq/cost} includes terms that regulate the ZMP modulation variables~$\bm{z}^{\text{ctrl}}_{0,i}$, $\bm{z}^{\text{ctrl}}_{T,i}$, the step location adjustment~$\bm{f}^{\text{ctrl}}_{T_{\text{new},i}}$, and the phase duration adjustment~$\delta T_{i}$, as well as a term that minimizes the DCM offset error~$\bm{b}^{\text{err}}_{T_{\text{new},i}}$. 
The weighting parameters are summarized in Table~\ref{tab:weights}, and the same values were used for all experiments.
The ZMP modulation cost terms contribute to preventing rapid changes in the desired contact wrenches. 
The step location and phase duration adjustment terms penalize deviations from the pre-planned footstep plan. 
By tuning their weights, the user can control how strongly the stepping strategy is exploited for balance recovery.
In addition, the DCM offset error term is assigned the largest weight, reflecting its dominant influence on walking stability. 
This cost structure allows the NMPC to flexibly use ZMP modulation, step location, and phase duration adjustments while prioritizing the minimization of the DCM offset.

The constraints consist of two equality constraints and three inequality constraints.
The first equality constraint \eqref{eq/ceq1} represents the dynamics constraints of the proposed NMPC.  
The first line of \eqref{eq/ceq1} enforces the nonlinear DCM error dynamics for the current support phase, while the second line of (\ref{eq/ceq1}) ensures the continuity of the error dynamics across the future previewed phases.  
Note that the DCM start point error of phase $i$ is equal to the DCM end point error of the previous phase as, $\bm{\xi}_{0,i}^{\text{err}} = \bm{\xi}_{T_{\text{new},i-1}}^{\text{err}} = \bm{f}_{T_{\text{new},i-1}}^{\text{ctrl}} + \bm{b}_{T_{\text{new},i-1}}^{\text{err}}. $
The second equality constraint~\eqref{eq/ceq2} ensures the continuity of the ZMP control input between adjacent phases such that the ZMP transition remains smooth across phase boundaries. 

The first inequality constraint~\eqref{eq/cineq1} defines the ZMP feasibility bounds. 
The ZMP control input at the start and end of each phase must remain within the local support foot. 
Thus, the bounds of the ZMP control inputs are determined by the foot size.
The second inequality constraint~\eqref{eq/cineq2} imposes limits on the step location adjustment. 
In the SSP, the robot must step within its reachable region. 
This leads to the following constraint:
$
\underline{\bm{f}}^{\text{ctrl}}_{T_{\text{new}},i,\text{SSP}} \leq 
\bm{f}^{\text{ctrl}}_{T_{\text{new}},i} \leq 
\overline{\bm{f}}^{\text{ctrl}}_{T_{\text{new}},i,\text{SSP}}.
$
During the DSP, no step location adjustments are allowed.
This leads to the following constraint: 
$
\bm{f}^{\text{ctrl}}_{T_{\text{new}},i} =~\bm{0}.
$
To accommodate the alternating SSP and DSP structures in the previewed phases, this constraint is formulated using the standard big-$M$ method~\cite{Conforti2014}. 
Here, $\bm{y}_j \in \mathbb{B}^2$ is a binary indicator vector: $\bm{y}_j = \bm{1}$ if phase $j$ is active, and $\bm{y}_j = \bm{0}$ otherwise. 
The vectors $\bm{0}$ and $\bm{1} \in \mathbb{R}^2$ consist of all zeros and ones, respectively. 
The variable $M$ is a sufficiently large constant that activates the corresponding constraints depending on the gait schedule.
The final inequality constraint~\eqref{eq/cineq3} imposes limits on the phase duration adjustment $\delta T_i$. 
These limits prevent an excessive increase in the swing foot speed during the SSP and impulsive ZMP shifts during the DSP.

The proposed NMPC problem is solved using Sequential Quadratic Programming (SQP)~\cite{Boggs1995SQP}, an iterative method for constrained nonlinear optimization.  
The solution is iteratively updated as, 
\begin{equation}
\bm{v}_{k+1} = \bm{v}_k + \beta_k \triangle \bm{v}_k,
\end{equation}
where $k$ denotes the iteration step and $\beta_k$ is determined via a line search algorithm in~\cite{Grandia2023NMPC}.  
To solve the nonlinear problem through multiple iterations, the active-set QP solver \texttt{qpOASES}~\cite{ferreau2014qpoases} is employed.  
To meet the requirements of real-time implementation, the number of SQP iterations is limited to 20, and the iteration is terminated early if the 2-norm of the SQP search direction $\triangle \bm{v}_k$ falls below $10^{-6}$.

\section{QP-based Whole-Body Control} \label{sec:wbc}

This section describes the WBC structure used to realize the outputs of the proposed NMPC and the walking pattern generator. 
Specifically, the WBC integrates tracking of the desired contact wrench obtained from the desired ZMP produced by the NMPC, as well as kinematic-level joint tracking based on the swing-foot trajectory from the NMPC and the CoM trajectory from the walking pattern generator.

First, the desired ZMP $\bm{z}^{\text{des}}_{t,1}$ generated by the NMPC is mapped to the desired contact wrench $\bm{w}_c^{\text{des}}$ for the support feet using the ZMP distribution method introduced in~\cite{kajita2010biped}. 
This step determines how the total ground reaction force should be allocated between the feet based on the desired ZMP location.

Next, to generate joint motions consistent with the desired CoM and foot trajectories, the desired joint angles of the lower body are computed using analytical inverse kinematics, while the desired joint angles of the upper body are fixed to their initial configuration.  
A kinematic-level PD controller is then used to compute the desired joint accelerations:
\begin{equation}
\ddot{\bm{q}}^{\text{des}} 
= \bm{K}_p^{\text{kin}} (\bm{q}^{\text{des}} - \bm{q})
- \bm{K}_d^{\text{kin}} \dot{\bm{q}},
\end{equation}
where $\bm{K}_p^{\text{kin}}$ and $\bm{K}_d^{\text{kin}}$ denote the kinematic-level PD gains.
These accelerations serve as references for the inverse-dynamics module.

To achieve both the desired joint accelerations and the desired contact wrench,
a QP-based WBC from~\cite{kim2020dynamicwbc} is employed.
Given $\ddot{\bm{q}}^{\text{des}}$ and $\bm{w}_c^{\text{des}}$, the WBC solves a QP
that incorporates the full-body humanoid dynamics and soft contact constraints,
and computes the feedforward torque command $\bm{\tau}^{\text{ff}}$.

Finally, the torque command applied to the robot actuators at 2~kHz is
\begin{equation}
\bm{\tau}^{\text{cmd}}
= \bm{\tau}^{\text{ff}}
+ \bm{K}_p (\bm{q}^{\text{des}} - \bm{q})
- \bm{K}_d \dot{\bm{q}},
\end{equation}
where $\bm{K}_p$ and $\bm{K}_d$ denote the motor-level joint-space PD gains.

\section{Results of Simulations and Experiments} 
\label{sec:result}
This section presents simulation and real-robot experiments to evaluate the proposed NMPC framework.
Simulations include ablation studies on different combinations of balance strategies, as well as comparative robustness evaluations against baseline controllers~\cite{egle2023step, kim2023foot, choe2023seamless} under external disturbances and on uneven and compliant terrains.
Real-world experiments validate the robot’s balance control behavior under external pushes and uneven terrain conditions.

\subsection{System overview}
The proposed method was validated through simulations in MuJoCo~\cite{todorov2012mujoco} and hardware experiments on the humanoid robot, TOCABI~\cite{schwartz2022design}.
TOCABI is a full-sized humanoid robot, measuring approximately 1.8 m in height and weighing about 100 kg. 
It comprises a total of 33 degrees of freedom: 16 in the arms, 12 in the legs, 3 in the waist, and 2 in the neck.
The proposed NMPC and walking pattern generator~\cite{wieber2006trajectory} were executed at 100 Hz, while the main control loop—responsible for ZMP trajectory planning, foot trajectory generation, and torque command synthesis—operated at 2 kHz.
The number of previewed phases $n_c$ in the NMPC was set to three to predict two upcoming walking steps, following the proposition in~\cite{zaytsv2015twostep}. 
As a QP solver, qpOASES~\cite{ferreau2014qpoases} was used to solve the QP optimization problems, including those for the walking pattern generator, WBC, and \eqref{eq/cost}, while RBDL~\cite{Felis2016} was employed to compute the kinematics and dynamics.
The reference durations of the SSP and DSP were set to 0.6 and 0.3~$s$, respectively.
The positive x-axis corresponds to the robot’s forward direction, and the positive z-axis points opposite to the gravity vector.
In real-world experiments, the proposed method was deployed on an onboard computer mounted on the back of the robot, equipped with a 3.7 GHz octa-core processor and 16 GB of RAM. 



\subsection{Simulations for comparative ablation study on balance strategy combinations}
The proposed NMPC framework integrates ZMP modulation, step location adjustment, and both SSP and DSP duration adjustments within a single optimization problem.
This section investigates whether such integration leads to meaningful improvements in robustness against external disturbances.
To this end, comparative simulations are conducted to analyze how different combinations of the aforementioned balance strategies influence balance performance under identical disturbance conditions.
The implementation of each strategy combination is summarized as follows:
\begin{itemize}
    \item Method 1) is the proposed approach, which combines the ZMP modulation, step location, step timing, and DSP duration adjustments (red line). 
    \item Method 2) includes the ZMP modulation, step location, and step timing adjustments with a fixed DSP duration (green line). 
    \item Method 3) includes ZMP modulation and step location adjustment with fixed SSP and DSP durations (blue line).  
    \item Method 4) includes only the ZMP modulation (black line). 
\end{itemize}
\begin{figure}[!t]
\centerline{\includegraphics[width=0.5\textwidth]{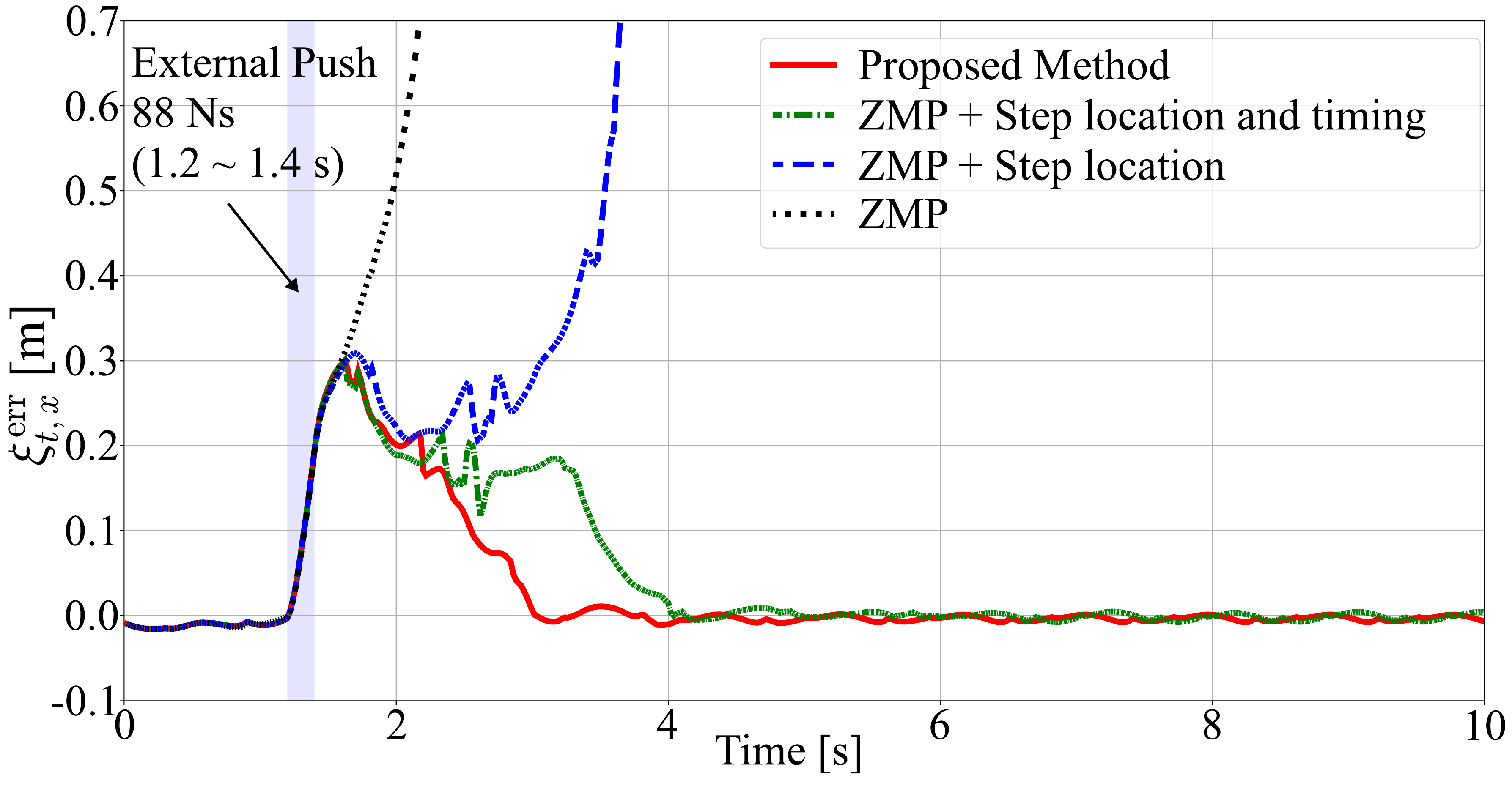}}
\caption{Comparative ablation results of the proposed NMPC under different combinations of balance strategies.}
\label{fig/sim_comparison}
\end{figure}
Fig.~\ref{fig/sim_comparison} shows the DCM error $\xi^{\text{err}}_{t,x}$ in the x-direction for each method. 
During the experiment, an external impulse of 88 $N\cdot s$ (corresponding to a 440 $N$ force applied for 0.2 $s$) was applied to the robot’s pelvis in the forward direction ($+x$) at 1.2 $s$ during in-place walking, as indicated by the light blue area in Fig.~\ref{fig/sim_comparison}. 
Method 4) results in the fastest fall since the ZMP control alone is insufficient to counteract the large disturbance.
Method 3) leads to a fall as the step timing cannot be adjusted in response to the disturbance. 
Although step location is optimized, the robot is required to wait for the predetermined step timing, causing the loss of balance to occur before the foot can be placed at the adjusted location.
In contrast, methods~1) and~2), which include phase duration adjustments, successfully maintain balance under the same disturbance. 
Notably, Method 1) exhibits faster disturbance rejection than Method 2).
By shortening the DSP duration, Method 1) transitions more quickly to the subsequent SSP, allowing the robot to perform more agile stepping and establish a new support polygon earlier.

\subsection{Simulations for comparative evaluation of three baseline controllers}
This section presents a comparative study evaluating the robustness of the proposed method against three baseline controllers~\cite{egle2023step,kim2023foot,choe2023seamless}.
To ensure fair comparisons, all other local controllers, such as the walking pattern generator and the WBC described in Sec.~\ref{sec:overview}, were kept identical across all methods. 
The reasons for selecting these three controllers as baselines are as follows. 
First, the proposed method, the LMPC method~\cite{egle2023step}, and the heuristic method~\cite{kim2023foot} all employ the same set of balance strategies—ZMP modulation, and adjustments of step location, step timing, and DSP duration.
Second, the NMPC-w/o-DSP method~\cite{choe2023seamless}, which incorporates all of the aforementioned strategies except for DSP duration adjustment, represents a state-of-the-art approach in this domain. 

The LMPC method~\cite{egle2023step} optimizes step location, step timing, and DSP duration based on linearized DCM–ZMP error dynamics, while ZMP modulation is handled separately by a DCM feedback controller.  
Specifically, the stepping strategy is activated only when the desired ZMP computed by the DCM feedback controller exceeds a predefined admissible set within the support polygon.  
This sequential use of balance strategies requires DCM error separation and projection to compensate for the resulting model mismatch.
The heuristic method~\cite{kim2023foot} employs a stepping controller with a DSP scaler that adjusts the DSP duration based on the additionally generated step location over a fixed nominal step location.  
The NMPC-w/o-DSP method~\cite{choe2023seamless} integrates ZMP modulation, step location, step timing, and time derivative of the centroidal angular momentum within a unified optimization framework, but computes these control inputs only during the SSP.  
For a fair comparison, the NMPC-w/o-DSP method was augmented with a DCM feedback controller during the DSP, and the DSP duration was kept fixed at its nominal value, since the original formulation cannot compute control inputs during the DSP. 
Moreover, the time derivative of the centroidal angular momentum was constrained to zero so that stepping performance could be compared fairly with the proposed method.
The same MPC weights and number of SQP iterations as those reported in~\cite{choe2023seamless} were used.

\subsubsection{Evaluation of robustness against external forces}

This section evaluates whether the proposed framework provides improved robustness against external forces compared to baseline controllers.
To this end, we assess the maximum external impulses that the robot can recover from under different disturbance conditions, focusing on both the disturbance direction and timing.

\begin{figure}[!t]
\centerline{\includegraphics[width=0.5\textwidth]{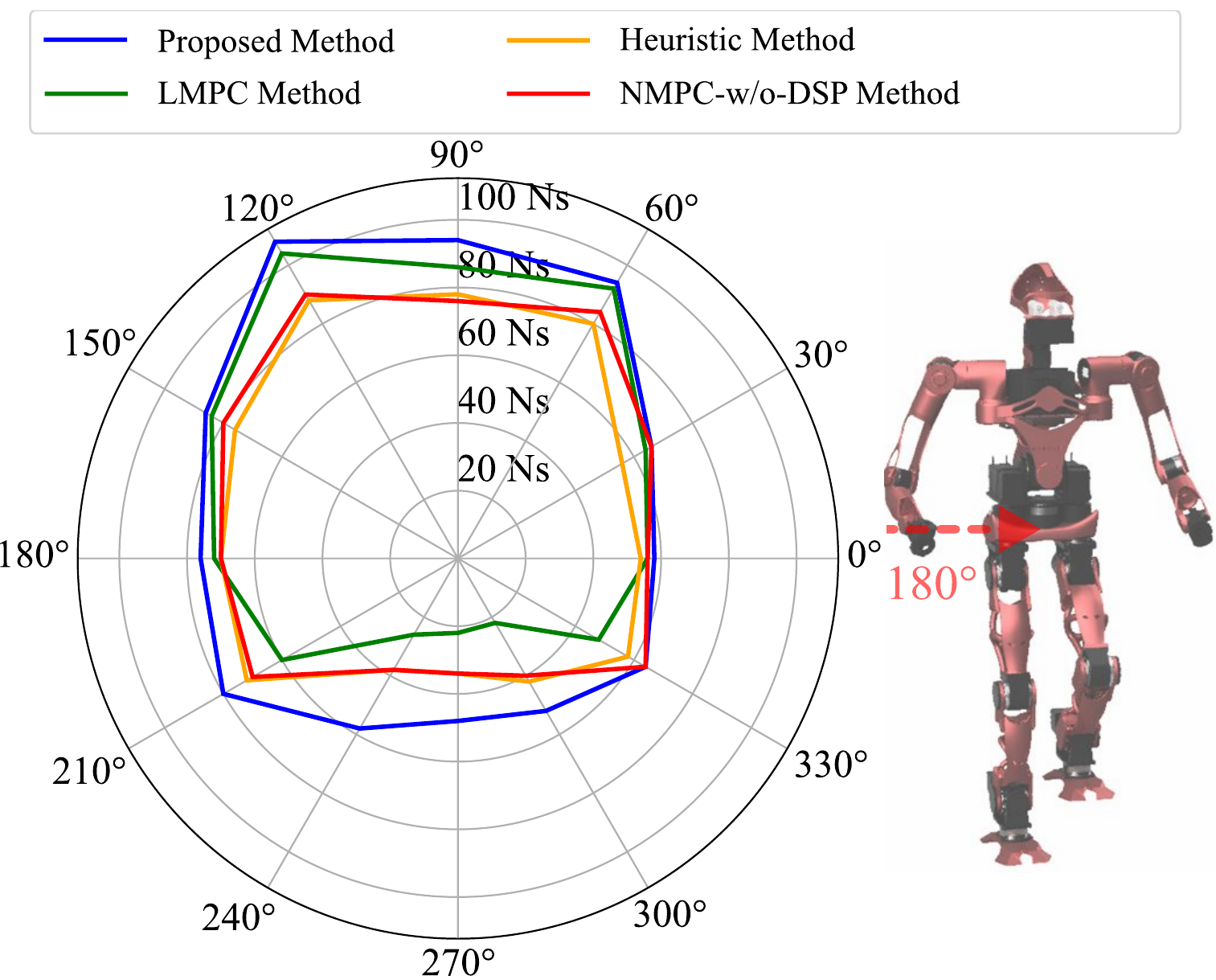}}
\caption{Comparison of maximum recoverable impulses from different push directions for the proposed method, the LMPC method~\cite{egle2023step}, the heuristic method~\cite{kim2023foot}, and NMPC-w/o-DSP method~\cite{choe2023seamless}.}
\label{fig/disturbance polygon}
\end{figure}
\textbf{1) Evaluation of maximum recoverable impulses from various directions:}  
Simulations were conducted to measure the maximum recoverable impulse applied from various directions.
This experiment aims to verify whether the proposed framework can consistently maintain balance under disturbances of arbitrary directions, which is a critical requirement for robust humanoid walking in real-world environments.
In each trial, an external push was applied to the robot’s pelvis for 0.2~$s$ during in-place walking, with the robot being in the SSP where the right foot served as the support foot and the left foot was in the swing phase.
The impulse magnitude was gradually increased until the robot failed to recover its balance.
Fig.~\ref{fig/disturbance polygon} presents the maximum impulse that the robot could withstand for each push direction. 

As shown in Fig.~\ref{fig/disturbance polygon}, the proposed method consistently achieved higher recoverable impulses than all three baselines across all directions.
The directions in which the robot could withstand the largest external disturbances were between 60° and 120°, where it tolerated impulses up to 108~$N\cdot s$, showing average improvements of 17.5\% over the heuristic method and 15.8\% over the NMPC-w/o-DSP method.
Even under backward disturbances (240°–300°), the proposed method maintained balance with impulses ranging from 48 to 58~$N\cdot s$, while all baselines failed to recover from impulses exceeding 40~$N\cdot s$.

\begin{figure}[!t]
\centerline{\includegraphics[width=0.55\textwidth]{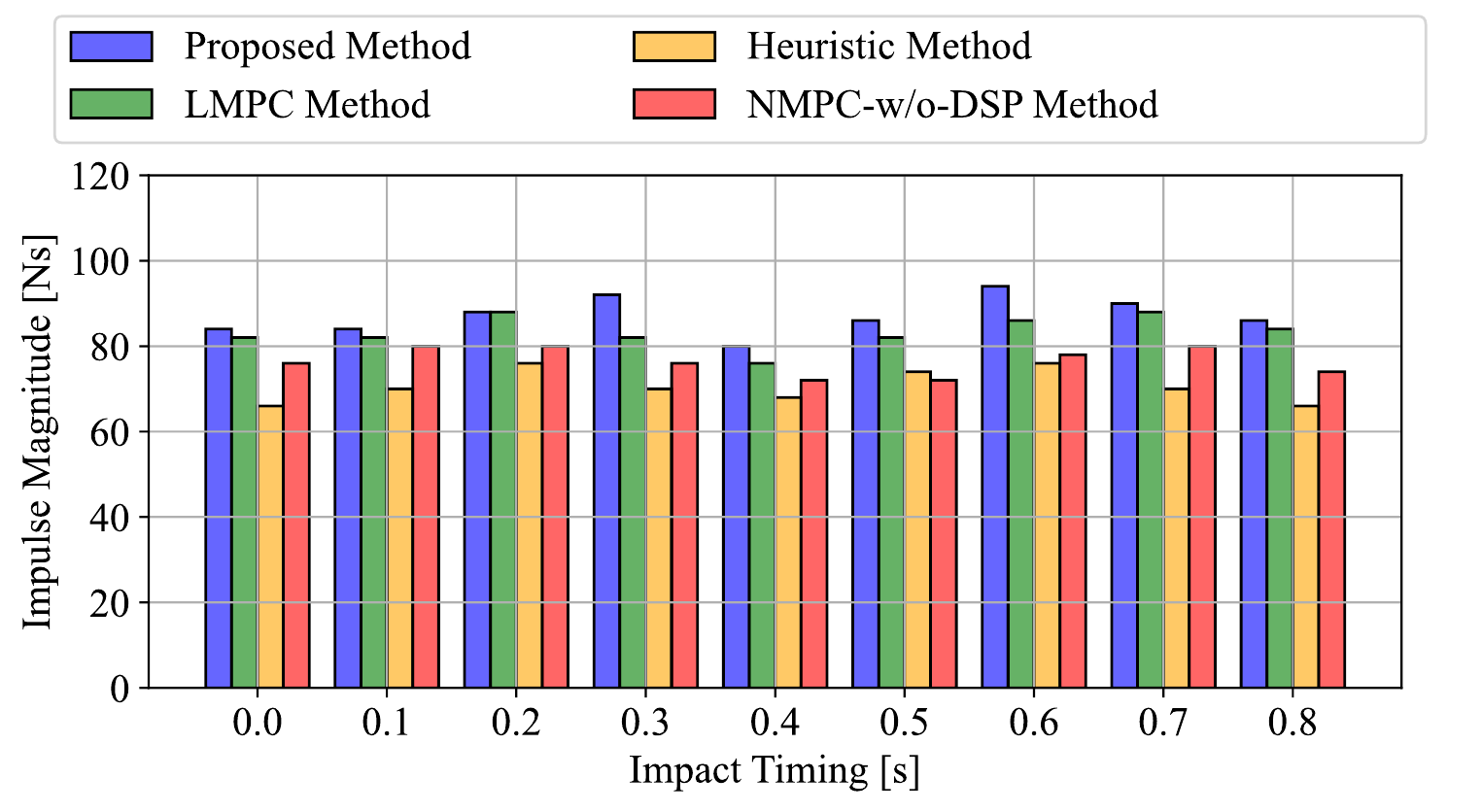}}
\caption{Comparative simulation of robustness to external impulse timing variations between the proposed method, the LMPC method~\cite{egle2023step}, the heuristic method~\cite{kim2023foot}, and the NMPC-w/o-DSP method~\cite{choe2023seamless}.
The graph represents the maximum impulse magnitude resistance in the forward direction.}
\label{fig/sim_coparison_timing}
\end{figure}

\textbf{2) Evaluation of maximum recoverable impulses at various timings:} 
Next, we evaluate robustness with respect to the timing of the external impulse within a single step.
This experiment aims to verify whether the proposed method can respond robustly to an external force regardless of when it occurs.
Such timing-invariant robustness is critical in practice, as external disturbances are unpredictable and may occur at any timing of locomotion.

Fig.~\ref{fig/sim_coparison_timing} shows the maximum impulse resistance in the forward ($+x$) direction, where external forces were applied for 0.2~$s$ at varying impulse timings within one step. 
The interval from 0 to 0.9 $s$ corresponds to a full step cycle, including both the SSP and DSP, and was discretized at 0.1 $s$ intervals to cover all representative gait phases.
The interval from 0 to 0.3~$s$ corresponds to the DSP, while the interval from 0.3 to 0.9~$s$ corresponds to the SSP. 
The external force magnitude was increased by 10~$N$ per trial until balance recovery failed. 
As shown in Fig.~\ref{fig/sim_coparison_timing}, the proposed method consistently achieved higher disturbance tolerance across all impact timings than the other baselines. 

\textbf{3) Discussion:} 
The performance differences observed in the disturbance direction and disturbance timing evaluations can be attributed to fundamental differences in controller architecture design.
First, the heuristic method~\cite{kim2023foot} lacks adaptability during the DSP and the late phase of the SSP due to its stepping mechanism.  
Specifically, step adjustment is deactivated at a predefined time before the end of the SSP to heuristically adjust the DSP duration.
As a result, step location and step timing cannot be effectively adjusted for disturbances occurring within this predefined time window during the SSP.
Moreover, when a disturbance occurs during the DSP, the controller is restricted to ZMP modulation alone.
Consequently, the controller becomes ineffective at responding to disturbances that occur near phase transitions or during the DSP.
Second, the LMPC method~\cite{egle2023step} activates step adjustments only when the DCM error cannot be corrected through ZMP modulation within the support polygon. This sequential activation scheme causes delays in deploying stepping strategies, reducing responsiveness to disturbances.  
Third, the NMPC-w/o-DSP method~\cite{choe2023seamless} exhibited reduced performance compared to the proposed method, as it relies solely on ZMP modulation via a DCM feedback controller during the DSP.
Since its nonlinear DCM–ZMP dynamics are formulated only for the SSP, the controller cannot exploit model-predictive balance control in the DSP, but instead depends on a simple DCM feedback strategy without the ability to adjust the DSP duration.
As a consequence, disturbances applied near phase transitions or during the DSP cannot be effectively compensated, leading to reduced robustness with respect to disturbance timing.

\subsubsection{Evaluation of robust walking on challenging terrain}
This section investigates whether the proposed NMPC can maintain stable walking performance on uneven or compliant terrain.
Humanoid robots in real-world environments often encounter uneven or compliant ground, where contact dynamics deviate from ideal rigid or flat assumptions.  
Evaluating the controller under such conditions is essential to verify its adaptability and robustness against variations in contact unevenness and compliance.  

\begin{table}[t]
\centering
\caption{Success rates under varying ground height variation settings.}
\renewcommand{\arraystretch}{1.2}
\setlength{\tabcolsep}{4pt}
\begin{tabular}{lccccc}
\toprule
\multirow{2}{*}{\textbf{Method}} & \multicolumn{5}{c}{\textbf{Maximum Elevation Point}} \\ 
\cmidrule(lr){2-6}
 & 0.0 & 0.01 & 0.02 & 0.03 & 0.04 \\ 
\midrule
\textbf{Proposed} & \textbf{5/5} & \textbf{5/5} & \textbf{5/5} & \textbf{4/5} & \textbf{2/5} \\
LMPC~\cite{egle2023step} & \textbf{5/5} & \textbf{5/5} & 2/5 & 0/5 & 0/5 \\
Heuristic~\cite{kim2023foot} & \textbf{5/5} & 4/5 & 2/5 & 0/5 & 0/5 \\
NMPC-w/o-DSP~\cite{choe2023seamless} & \textbf{5/5} & \textbf{5/5} & 3/5 & 0/5 & 0/5 \\
\bottomrule
\end{tabular}
\label{tab:succ_hfield}
\end{table}

\textbf{1) Simulation Terrain Setting:}  
To simulate uneven terrain, MuJoCo’s \texttt{hfield} geom was used to represent the ground as a height map.
A randomly sampled grayscale image was used to define the spatial distribution of ground height variations.
The overall level of unevenness was controlled by the vertical scaling factor \texttt{elevation\_z} of the \texttt{hfield} size parameter, which determines the maximum elevation difference.
By increasing \texttt{elevation\_z} while keeping the horizontal dimensions fixed, terrain profiles with progressively larger height variations were generated.

To emulate compliant terrain, the \texttt{timeconst} parameter in the \texttt{solref} setting of the ground \texttt{geoms} was increased.
In MuJoCo, a larger \texttt{timeconst} results in slower constraint resolution, leading to increased contact penetration between the foot and the ground.
This setting provides a simplified approximation of compliant ground behavior.
\begin{figure*}[!t]
    \centering
    \includegraphics[width=\textwidth]{terrain_snapshot.pdf}
    \caption{Snapshots of the humanoid robot walking on uneven terrain (top) and compliant terrain (bottom) in MuJoCo. In these snapshots, the \texttt{elevation\_z} parameter for the uneven terrain experiment was set to 0.03, and the \texttt{timeconst} parameter for the compliant terrain experiment was set to 0.20.
}
    \label{fig:terrain_snapshot}
\end{figure*}

\begin{table}[t]
\centering
\caption{Success rates under varying ground compliance settings.}
\renewcommand{\arraystretch}{1.2}
\setlength{\tabcolsep}{4pt}
\begin{tabular}{lccccc}
\toprule
\multirow{2}{*}{\textbf{Method}} & 
\multicolumn{5}{c}{\textbf{Time Constant}} \\
\cmidrule(lr){2-6}
 & 0.16 & 0.17 & 0.18 & 0.19 & 0.20 \\ 
\midrule
\textbf{Proposed} & \textbf{5/5} & \textbf{5/5} & \textbf{5/5} & \textbf{5/5} & \textbf{5/5} \\
LMPC~\cite{egle2023step} & \textbf{5/5} & \textbf{5/5} & 3/5 & 1/5 & 0/5 \\
Heuristic~\cite{kim2023foot} & \textbf{5/5} & \textbf{5/5} & \textbf{5/5} & 4/5 & 2/5 \\
NMPC-w/o-DSP~\cite{choe2023seamless} & \textbf{5/5} & \textbf{5/5} & \textbf{5/5} & \textbf{5/5} & \textbf{5/5} \\
\bottomrule
\end{tabular}
\label{tab:succ_timeconst}
\end{table}

\textbf{2) Results:} 
Fig.~\ref{fig:terrain_snapshot} presents snapshots of the walking experiments conducted on uneven and compliant terrains in MuJoCo, while Tables~\ref{tab:succ_hfield} and \ref{tab:succ_timeconst} report the corresponding success rates, respectively.
Walking success is defined as the humanoid robot successfully following the nominal gait pattern, which consists of moving forward 2~$m$ with a 0.3~$m$ step stride, without falling. 

On uneven terrain experiments, the proposed method consistently outperformed all baseline controllers across all tested levels of ground height variation, sustaining reliable walking even as the terrain became more irregular.
In contrast, all baseline controllers exhibited a noticeable degradation in performance once the maximum elevation exceeded approximately 0.02~$m$.

On compliant terrain experiments, the proposed and the NMPC-w/o-DSP methods succeeded in all trials over the entire tested range of \texttt{timeconst} values.
However, the LMPC and heuristic methods became progressively less reliable as the ground softness increased, with frequent failures observed at \texttt{timeconst} values of 0.18 and above.

\textbf{3) Discussion:}
The improved robustness of the proposed method on both uneven and compliant terrains can be attributed to the seamless integration of multiple balance strategies within a unified NMPC framework.
As mentioned in~\cite{kim2025tro}, during walking, small DCM errors induced by foot–ground impacts, modeling uncertainties, or mild terrain irregularities can be effectively reduced by ZMP modulation alone.
However, when larger DCM errors arise due to stronger external pushes or more severe terrain irregularities, the controllability afforded by ZMP modulation becomes insufficient, and additional strategies such as step adjustments are required to prevent loss of balance.
In such cases, as discussed in~\cite{choe2023seamless}, seamless integration of balance strategies can offer clear advantages over sequential and heuristic approaches.

In the LMPC method, stepping is typically triggered when the ZMP command required to reduce the DCM error exceeds the support polygon.
In scenarios with persistent disturbances caused by uneven or compliant ground, this sequential mechanism can cause stepping control to intervene irregularly in the overall balance control.
In contrast, the proposed method computes all balance strategies within a single optimization problem based on the current DCM error, allowing these strategies to be utilized in a coordinated manner.
Consistent with this interpretation, the proposed method and the NMPC-w/o-DSP method exhibited robust performance on compliant terrain.
Furthermore, the proposed method seamlessly incorporates ZMP modulation and duration adjustment during the DSP, which contributes to the additional performance gains compared to the NMPC-w/o-DSP method, as observed in the uneven terrain experiments.
\begin{figure*}[!t]
    \centering
    \includegraphics[width=1.0\textwidth]{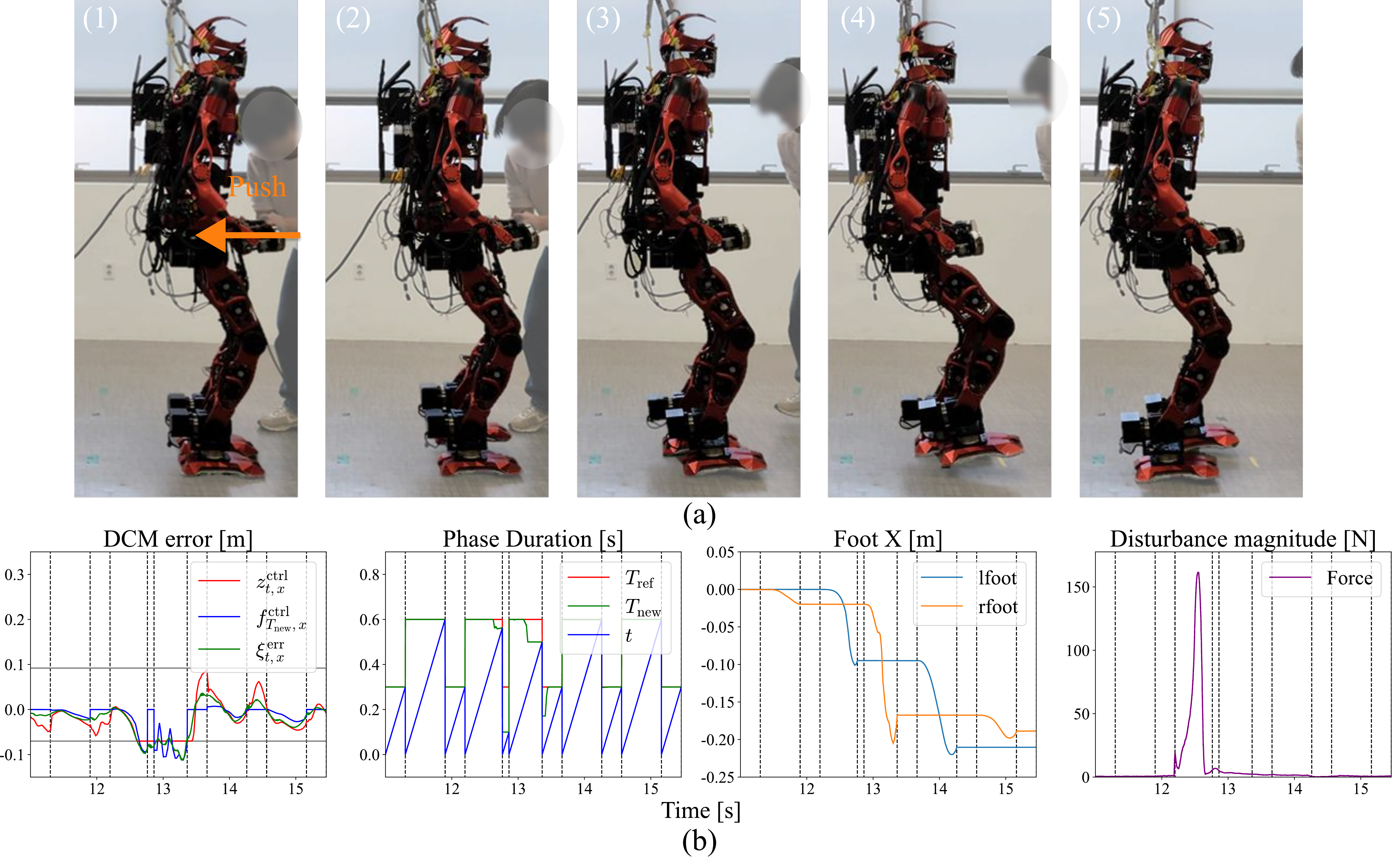}
    \caption{Experimental results of the proposed method under backward external disturbance.
    Snapshots of the walking experiment under a backward push are shown in (a).
    The corresponding output data are shown in (b), including the DCM errors $\bm{\xi}_{t,x}^{\text{err}}$, ZMP modulation $\bm{z}_{t,x}^{\text{ctrl}}$, and step location adjustment $\bm{f}_{T_{\text{new}},x}^{\text{ctrl}}$ along the $x$-direction, as well as the optimized phase durations $T_{\text{new}}$, backward foot trajectory, and disturbance magnitudes measured by the force/torque sensor.}
    \label{fig:real_robot_x}
\end{figure*}

\begin{figure*}[!t]
    \centering
    \includegraphics[width=1.0\textwidth]{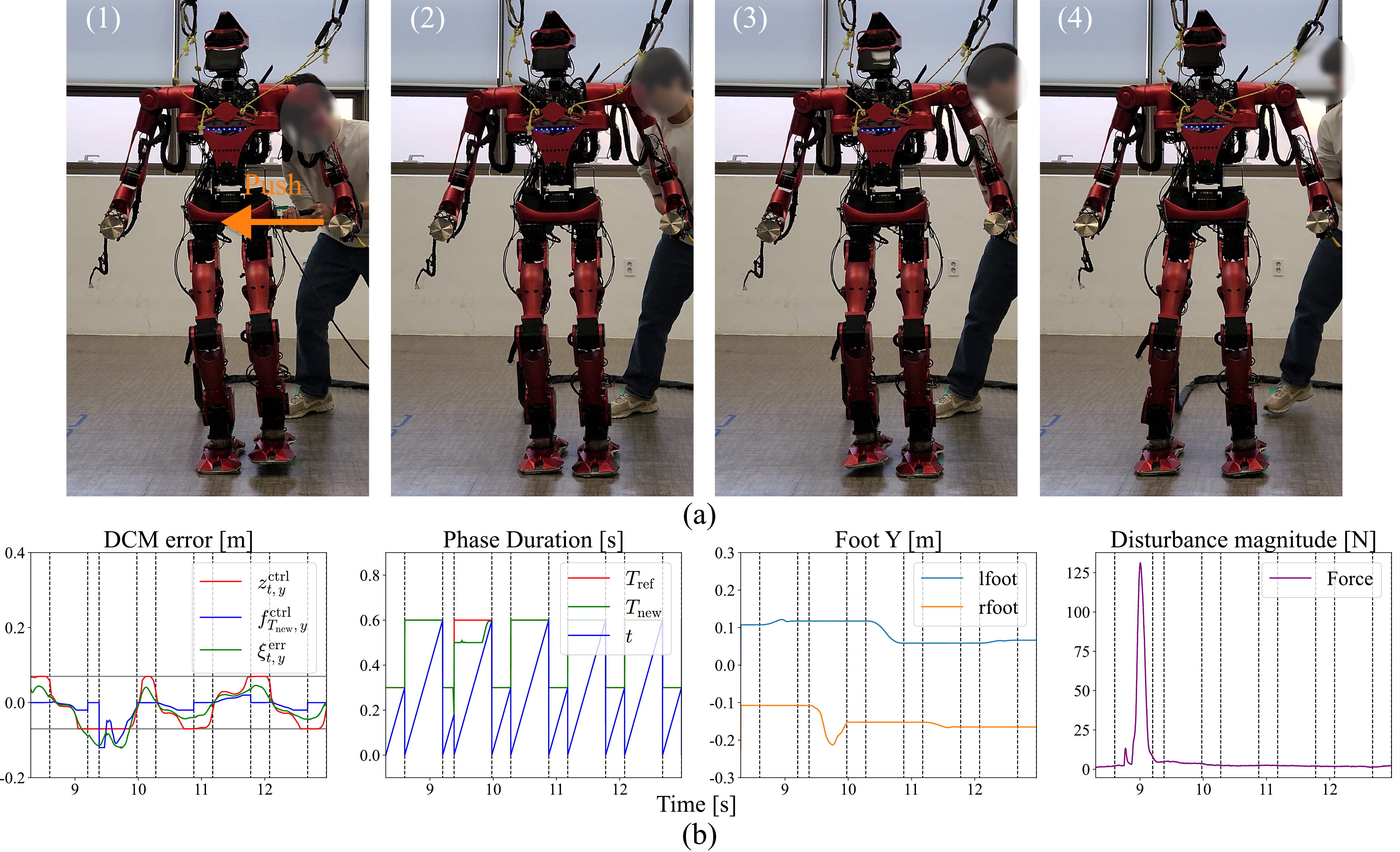}
    \caption{Experimental results of the proposed method under lateral external disturbance.
    Snapshots of the walking experiment under a lateral push are shown in (a).
    The corresponding output data are shown in (b), including the DCM errors $\bm{\xi}_{t,y}^{\text{err}}$, ZMP modulation $\bm{z}_{t,y}^{\text{ctrl}}$, and step location adjustment $\bm{f}_{T_{\text{new}},y}^{\text{ctrl}}$ along the $y$-direction, as well as the optimized phase durations $T_{\text{new}}$, lateral foot trajectory, and disturbance magnitudes measured by the force/torque sensor.}
    \label{fig:real_robot_y}
\end{figure*}

\subsection{Hardware experiments}
{
In this section, the experimental results on a real robot are presented to evaluate the robustness of the proposed method.

\subsubsection{Evaluation of robust walking under external push}
Real-robot experiments were conducted in which the humanoid robot was subjected to external forces in both the $x$- and $y$-directions while walking in place.
To assess the performance of the proposed method, physical pushes were delivered to the pelvis using a tool instrumented with a force/torque (F/T) sensor, and the resulting impulse magnitude was measured from this sensor.
All controller parameters were kept identical to those used in the simulation experiments, except for the motor-level PD gains.

In the backward push experiment, the robot was subjected to an external push of 26.98~$N\cdot s$.
Fig.~\ref{fig:real_robot_x}(a) and (b) show the experimental snapshots and the corresponding data, respectively.
The snapshot illustrates that the robot successfully maintains balance, overcoming the disturbance through ZMP modulation together with adjustments of step location, step timing, and DSP duration.
At approximately 12.2~$s$, a backward external force is applied to the robot pelvis, which generates a DCM error $\xi_{t,x}^{\text{err}}$ (green line in the first column) in the negative $x$-direction.
In response, the proposed NMPC simultaneously produces a ZMP modulation command $z_{t,x}^{\text{ctrl}}$ (red line in the first column) and a step location adjustment $f_{T_{\text{new}},x}^{\text{ctrl}}$ (blue line in the first column) to reduce the DCM error.
The ZMP command quickly saturates at the ZMP constraint of $-7$~$cm$ around 12.6~$s$, after which the proposed NMPC further decreases the DCM error by shortening the step timing $T_{\text{new}}$ (green line in the second column) from the nominal SSP duration of 0.60~$s$ to 0.56~$s$.
In the subsequent DSP, ZMP modulation is exploited up to the constraint, and the DSP duration $T_{\text{new}}$ is additionally reduced to its lower bound of 0.10~$s$.
As a result, the robot performs three backward steps to maintain balance and returns to the nominal walking pattern at around 13.7~$s$, approximately 1.5~$s$ after the disturbance, having moved backward by about 16.7~$cm$.

In the lateral push experiment, the robot was subjected to an external push of 21.66~$N\cdot s$.
Fig.~\ref{fig:real_robot_y}(a) and (b) show the experimental snapshots and the corresponding data, respectively.
At approximately 8.7~$s$, a lateral external force applied to the support foot generates a DCM error $\xi_{t,y}^{\text{err}}$ in the negative $y$-direction.
In response, the proposed NMPC simultaneously produces a ZMP modulation command $z_{t,y}^{\text{ctrl}}$ and a step location adjustment $f_{T_{\text{new}},y}^{\text{ctrl}}$ to reduce the DCM error.
The ZMP command quickly saturates at the ZMP constraint of $-7$~$cm$, and the step-location adjustment is generated up to $-9$~$cm$.
In the subsequent DSP, ZMP modulation is exploited up to the constraint, and the DSP duration $T_{\text{new}}$ is further reduced to 0.18~$s$.
During the following SSP, the NMPC simultaneously optimizes the ZMP modulation, step location, and step timing to minimize the DCM error.
As a result, the robot performs two lateral steps to maintain balance and returns to the nominal walking pattern at around 10.0~$s$, approximately 1.3~$s$ after the disturbance, having moved laterally by about 5.9~$cm$.

As demonstrated by the experimental results, the proposed method achieves coordinated balance control during the SSP by actively exploiting step adjustments even when the ZMP command has not yet reached its constraint boundary.
This behavior contrasts with the LMPC method~\cite{egle2023step}, which relies on a sequential activation of the stepping strategy.
Moreover, during the DSP, the proposed method enables dynamically consistent ZMP modulation and duration adjustment by leveraging the proposed nonlinear DCM–ZMP dynamics, the ZMP-input continuity constraint, and the big-$M$ footstep constraints.
As a result, the proposed NMPC can compute dynamically feasible balance-control inputs regardless of whether the robot is in SSP or DSP.
This result contrasts with the NMPC-w/o-DSP method~\cite{choe2023seamless}, which can handle the nonlinear DCM–ZMP dynamics only during the SSP.
}

\begin{figure*}[!t]
    \centering
    \includegraphics[width=\textwidth]{real_robot_obs.pdf}
    \caption{Snapshots of the robot momentarily stepping on a mat during walking, simulating an uneven terrain condition.}
    \label{fig:real_robot_obs}
\end{figure*}

\subsubsection{Evaluation of robust walking under unexpected terrain perturbations}
To further evaluate the robustness of the proposed method under unexpected terrain perturbations, a real-robot experiment was conducted under uneven ground conditions.
This experiment aims to assess whether the proposed method can maintain stable walking when sudden and unmodeled changes in ground height occur at the swing-foot landing location, without relying on exteroceptive terrain sensing.
Fig.~\ref{fig:real_robot_obs} shows snapshots of the uneven terrain experiment.
While the robot was walking in place, a mat with a height of 3~$cm$ was placed at the upcoming landing location of the swing foot, creating an abrupt and unforeseen change in the contact condition.
Despite the absence of exteroceptive sensing or ankle force/torque feedback, the robot successfully maintained its balance by exploiting a coordinated combination of ankle and stepping strategies enabled by the proposed NMPC and the QP-based WBC.

\section{Conclusion} \label{sec:conclusion}
This paper proposes a phase-based NMPC framework for robust humanoid balance control.  
The proposed controller simultaneously optimizes ZMP modulation, step location, step timing (SSP duration), and DSP duration, enabling the seamless integration of multiple balance strategies.
This unified formulation enables dynamically consistent and phase-invariant balance control, allowing different strategies to be exploited in a coordinated manner.
Through extensive simulation and real-robot experiments across diverse disturbance and terrain scenarios, it is demonstrated that this coordinated use of balance strategies significantly enhances robustness.  
In particular, the proposed method consistently outperformed baseline methods, achieving notable performance improvements across all evaluated experimental scenarios.

As future work, the proposed framework will be extended by integrating reinforcement learning with the existing model predictive control structure, following recent efforts that combine model-based control with learning-based policies~\cite{Kang2023RLPlus, Liu2025Opt2Skill, Jeon2025ResidualMPC}.
Learning-based components will be explored as a complementary layer to the proposed phase-based NMPC, providing adaptive corrections for model mismatches, contact uncertainties, and environmental properties such as friction that are difficult to model explicitly.
By retaining the interpretability, stability guarantees, and constraint-handling capabilities of the model-based controller while augmenting it with data-driven adaptability, such a hybrid approach is expected to further enhance robustness in highly uncertain and unstructured real-world environments.
\section*{DECLARATIONS}

\subsection*{Conflict of Interest}
The authors declare that there is no competing financial interest or personal relationship that could have appeared to influence the work reported in this paper.

\subsection*{Authors' Contributions}
Conceptualization, methodology, formal analysis: Kwanwoo Lee; 
Software: Kwanwoo Lee; 
Validation: Kwanwoo Lee, Gyeongjae Park; 
Investigation: Kwanwoo Lee, Gyeongjae Park; 
Resources: Jaeheung Park; 
Data curation: Kwanwoo Lee; 
Writing—original draft: Kwanwoo Lee;
Writing—review and editing: Myeong-Ju Kim; 
Supervision: Jaeheung Park; 
Project administration: Jaeheung Park;
Funding acquisition: Jaeheung Park

\subsection*{Funding }
This work was supported by the National Research Foundation of Korea(NRF) grant funded by the Korea government(MSIT) (No. 2021R1A2C3005914).


\clearafterbiography
\relax 


\begin{reference}
\bibitem{kuindersma2016optimization} S. Kuindersma, R. Deits, M. Fallon, A. Valenzuela, H. Dai, F. Permenter, T. Koolen, P. Marion, and R. Tedrake, ``Optimization-based locomotion planning, estimation, and control design for the Atlas humanoid robot,'' {\it Autonomous Robots}, vol. 40, no. 3, pp. 429--455, 2016.

\bibitem{khadiv2020walking} M. Khadiv, A. Herzog, S. A. A. Moosavian, and L. Righetti, ``Walking control based on step timing adaptation,'' {\it IEEE Transactions on Robotics}, vol. 36, no. 3, pp. 629--643, 2020.

\bibitem{jeong2019robust} H. Jeong, I. Lee, J. Oh, K. K. Lee, and J.-H. Oh, ``A robust walking controller based on online optimization of ankle, hip, and stepping strategies,'' {\it IEEE Transactions on Robotics}, vol. 35, no. 6, pp. 1367--1386, 2019.

\bibitem{shafiee2019online} M. Shafiee, G. Romualdi, S. Dafarra, F. J. A. Chavez, and D. Pucci, ``Online DCM trajectory generation for push recovery of torque-controlled humanoid robots,'' in {\it 2019 IEEE-RAS 19th International Conference on Humanoid Robots (Humanoids)}, pp. 671--678, 2019.

\bibitem{griffin2017walking} R. J. Griffin, G. Wiedebach, S. Bertrand, A. Leonessa, and J. Pratt, ``Walking stabilization using step timing and location adjustment on the humanoid robot, Atlas,'' in {\it 2017 IEEE/RSJ International Conference on Intelligent Robots and Systems (IROS)}, pp. 667--673, 2017.

\bibitem{ding2022oampc} Y. Ding, C. Khazoom, M. Chignoli, and S. Kim, ``Orientation-aware model predictive control with footstep adaptation for dynamic humanoid walking,'' in {\it 2022 IEEE-RAS 21st International Conference on Humanoid Robots (Humanoids)}, pp. 299--305, 2022.

\bibitem{choe2023seamless} J. Choe, J.-H. Kim, S. Hong, J. Lee, and H.-W. Park, ``Seamless reaction strategy for bipedal locomotion exploiting real-time nonlinear model predictive control,'' {\it IEEE Robotics and Automation Letters}, vol. 8, no. 8, pp. 5031--5038, 2023.

\bibitem{scianca2020stability} N. Scianca, D. De Simone, L. Lanari, and G. Oriolo, ``MPC for humanoid gait generation: Stability and feasibility,'' {\it IEEE Transactions on Robotics}, vol. 36, no. 4, pp. 1171--1188, 2020.

\bibitem{hong2020real} S. Hong, J.-H. Kim, and H.-W. Park, ``Real-time constrained nonlinear model predictive control on SO(3) for dynamic legged locomotion,'' in {\it 2020 IEEE/RSJ International Conference on Intelligent Robots and Systems (IROS)}, pp. 3982--3989, 2020.

\bibitem{kim2025real} J. Kim, H. Lee, and J. Park, ``Real-time whole-body model predictive control for bipedal locomotion with novel kino-dynamic model and warm-start method,'' {\it International Journal of Control, Automation and Systems}, vol. 23, no. 11, pp. 3338--3348, 2025.

\bibitem{kim2025tro} M.-J. Kim, D. Lim, G. Park, K. Lee, and J. Park, ``A model predictive capture point control framework for robust humanoid balancing via ankle, hip, and stepping strategies,'' {\it IEEE Transactions on Robotics}, vol. 41, pp. 3297--3316, 2025.

\bibitem{kajita20013d} S. Kajita, F. Kanehiro, K. Kaneko, K. Yokoi, and H. Hirukawa, ``The 3D linear inverted pendulum mode: A simple modeling for a biped walking pattern generation,'' in {\it Proc. 2001 IEEE/RSJ International Conference on Intelligent Robots and Systems (IROS)}, vol. 1, pp. 239--246, 2001.

\bibitem{carlo2018srbmpc} J. Di Carlo, P. M. Wensing, B. Katz, G. Bledt, and S. Kim, ``Dynamic locomotion in the MIT Cheetah 3 through convex model-predictive control,'' in {\it 2018 IEEE/RSJ International Conference on Intelligent Robots and Systems (IROS)}, pp. 1--9, 2018.

\bibitem{takenaka2009real} T. Takenaka, T. Matsumoto, and T. Yoshiike, ``Real time motion generation and control for biped robot-1st report: Walking gait pattern generation,'' in {\it 2009 IEEE/RSJ International Conference on Intelligent Robots and Systems (IROS)}, pp. 1084--1091, 2009.

\bibitem{englsberger2015three} J. Englsberger, C. Ott, and A. Albu-Sch{\"a}ffer, ``Three-dimensional bipedal walking control based on divergent component of motion,'' {\it IEEE Transactions on Robotics}, vol. 31, no. 2, pp. 355--368, 2015.

\bibitem{vukobratovic2004zero} M. Vukobratovi{\'c} and B. Borovac, ``Zero-moment point---thirty five years of its life,'' {\it International Journal of Humanoid Robotics}, vol. 1, no. 1, pp. 157--173, 2004.

\bibitem{griffin2023reachability} R. Griffin, J. Foster, S. Pasano, B. Shrewsbury, and S. Bertrand, ``Reachability aware capture regions with time adjustment and cross-over for step recovery,'' in {\it 2023 IEEE-RAS 22nd International Conference on Humanoid Robots (Humanoids)}, pp. 1--8, 2023.

\bibitem{kim2023foot} M.-J. Kim, D. Lim, G. Park, and J. Park, ``Foot stepping algorithm of humanoids with double support time adjustment based on capture point control,'' in {\it 2023 IEEE International Conference on Robotics and Automation (ICRA)}, pp. 12198--12204, 2023.

\bibitem{egle2023step} T. Egle, J. Englsberger, and C. Ott, ``Step and timing adaptation during online DCM trajectory generation for robust humanoid walking with double support phases,'' in {\it 2023 IEEE-RAS 22nd International Conference on Humanoid Robots (Humanoids)}, pp. 1--8, 2023.

\bibitem{englsberger2017smooth} J. Englsberger, G. Mesesan, and C. Ott, ``Smooth trajectory generation and push-recovery based on divergent component of motion,'' in {\it 2017 IEEE/RSJ International Conference on Intelligent Robots and Systems (IROS)}, pp. 4560--4567, 2017.

\bibitem{koolen2012capturability} T. Koolen, T. de Boer, J. Rebula, A. Goswami, and J. Pratt, ``Capturability-based analysis and control of legged locomotion, Part 1: Theory and application to three simple gait models,'' {\it The International Journal of Robotics Research}, vol. 31, no. 9, pp. 1094--1113, 2012.

\bibitem{wieber2006trajectory} P.-B. Wieber, ``Trajectory free linear model predictive control for stable walking in the presence of strong perturbations,'' in {\it 2006 6th IEEE-RAS International Conference on Humanoid Robots}, pp. 137--142, 2006.

\bibitem{kajita2010biped} S. Kajita {\it et al.}, ``Biped walking stabilization based on linear inverted pendulum tracking,'' in {\it 2010 IEEE/RSJ International Conference on Intelligent Robots and Systems (IROS)}, pp. 4489--4496, 2010.

\bibitem{kim2020dynamicwbc} D. Kim, S. J. Jorgensen, J. Lee, J. Ahn, J. Luo, and L. Sentis, ``Dynamic locomotion for passive-ankle biped robots and humanoids using whole-body locomotion control,'' {\it The International Journal of Robotics Research}, vol. 39, no. 8, pp. 936--956, 2020.

\bibitem{Conforti2014} M. Conforti, G. Cornu{\'e}jols, and G. Zambelli, ``Integer programming models,'' in {\it Integer Programming}, pp. 45--84, Springer, Cham, 2014.

\bibitem{Boggs1995SQP} P. T. Boggs and J. W. Tolle, ``Sequential quadratic programming,'' {\it Acta Numerica}, vol. 4, pp. 1--51, 1995.

\bibitem{Grandia2023NMPC} R. Grandia, F. Jenelten, S. Yang, F. Farshidian, and M. Hutter, ``Perceptive locomotion through nonlinear model-predictive control,'' {\it IEEE Transactions on Robotics}, vol. 39, no. 5, pp. 3402--3421, 2023.

\bibitem{ferreau2014qpoases} H. J. Ferreau, C. Kirches, A. Potschka, H. G. Bock, and M. Diehl, ``qpOASES: A parametric active-set algorithm for quadratic programming,'' {\it Mathematical Programming Computation}, vol. 6, no. 4, pp. 327--363, 2014.

\bibitem{todorov2012mujoco} E. Todorov, T. Erez, and Y. Tassa, ``MuJoCo: A physics engine for model-based control,'' in {\it 2012 IEEE/RSJ International Conference on Intelligent Robots and Systems (IROS)}, pp. 5026--5033, 2012.

\bibitem{schwartz2022design} M. Schwartz, J. Sim, J. Ahn, S. Hwang, Y. Lee, and J. Park, ``Design of the humanoid robot TOCABI,'' in {\it 2022 IEEE-RAS 21st International Conference on Humanoid Robots (Humanoids)}, pp. 322--329, 2022.

\bibitem{zaytsv2015twostep} P. Zaytsev, S. J. Hasaneini, and A. Ruina, ``Two steps is enough: No need to plan far ahead for walking balance,'' in {\it 2015 IEEE International Conference on Robotics and Automation (ICRA)}, pp. 6295--6300, 2015.

\bibitem{Felis2016} M. L. Felis, ``RBDL: An efficient rigid-body dynamics library using recursive algorithms,'' {\it Autonomous Robots}, vol. 41, no. 2, pp. 495--511, 2017.

\bibitem{Kang2023RLPlus} D. Kang, J. Cheng, M. Zamora, F. Zargarbashi, and S. Coros, ``RL+ model-based control: Using on-demand optimal control to learn versatile legged locomotion,'' {\it IEEE Robotics and Automation Letters}, vol. 8, no. 10, pp. 6619--6626, 2023.

\bibitem{Liu2025Opt2Skill} F. Liu, Z. Gu, Y. Cai, Z. Zhou, H. Jung, J. Jang, S. Zhao, S. Ha, Y. Chen, and D. Xu, ``Opt2Skill: Imitating dynamically-feasible whole-body trajectories for versatile humanoid loco-manipulation,'' {\it IEEE Robotics and Automation Letters}, vol. 10, no. 11, pp. 12261--12268, 2025.

\bibitem{Jeon2025ResidualMPC} S. H. Jeon, H. J. Lee, S. Hong, and S. Kim, ``Residual MPC: Blending reinforcement learning with GPU-parallelized model predictive control,'' {\it arXiv preprint arXiv:2510.12717}, 2025.
\end{reference}
\end{document}